\documentclass[graybox]{svmult}

\usepackage{mathptmx}       
\usepackage{helvet}         
\usepackage{courier}        
\usepackage{type1cm}        
\usepackage{makeidx}         
\usepackage[pdftex]{graphicx}        
\usepackage[pdftex]{xcolor}
\usepackage{multicol}        
\usepackage[bottom]{footmisc}
\usepackage{amsmath}

\newcommand{\ky}[1]{\textcolor{black}{#1}}

\makeindex             

\begin{document}

\title*{Caption Generation of Robot Behaviors based on Unsupervised Learning of Action Segments}
\author{Koichiro Yoshino$^{1,2}$, Kohei Wakimoto$^1$, Yuta Nishimura$^1$, Satoshi Nakamura$^{1,2}$}
\institute{$^1$Nara Institute of Science and Technology \at 8916-5, Takayama, Ikoma, Nara, 6300192, Japan \email{koichiro at is.naist.jp}
\and $^2$Center for Advanced Intelligence Project (AIP), RIKEN \at }
%
%
\maketitle

\abstract*{Bridging robot action sequences and their natural language captions is an important task to increase explainability of human assisting robots in their recently evolving field.
In this paper, we propose a system for generating natural language captions that describe behaviors of human assisting robots.
The system describes robot actions by using robot observations; histories from actuator systems and cameras, toward end-to-end bridging between robot actions and natural language captions.
Two reasons make it challenging to apply existing sequence-to-sequence models to this mapping:
1) it is hard to prepare a large-scale dataset for any kind of robots and their environment, and
2) there is a gap between the number of samples obtained from robot action observations and generated word sequences of captions. 
We introduced unsupervised segmentation based on K-means clustering
to unify typical robot observation patterns into a class.
This method makes it possible for the network to learn the relationship
from a small amount of data.  
Moreover, we utilized a chunking method based on byte-pair encoding
(BPE) to fill in the gap between the number of samples of robot
action observations and words in a caption.
We also applied an attention mechanism to the segmentation task.
Experimental results show that the proposed model based on unsupervised learning can generate better descriptions than other methods.
We also show that the attention mechanism did not work well in our low-resource setting.}

\abstract{Bridging robot action sequences and their natural language captions is an important task to increase explainability of human assisting robots in their recently evolving field.
In this paper, we propose a system for generating natural language captions that describe behaviors of human assisting robots.
The system describes robot actions by using robot observations; histories from actuator systems and cameras, toward end-to-end bridging between robot actions and natural language captions.
Two reasons make it challenging to apply existing sequence-to-sequence models to this mapping:
1) it is hard to prepare a large-scale dataset for any kinds of robots and their environment, and
2) there is a gap between the number of samples obtained from robot action observations and generated word sequences of captions. 
We introduced unsupervised segmentation based on K-means clustering
to unify typical robot observation patterns into a class.
This method makes it possible for the network to learn the relationship
from a small amount of data.  
Moreover, we utilized a chunking method based on byte-pair encoding
(BPE) to fill in the gap between the number of samples of robot
action observations and words in a caption.
We also applied an attention mechanism to the segmentation task.
Experimental results show that the proposed model based on unsupervised learning can generate better descriptions than other methods.
We also show that the attention mechanism did not work well in our low-resource setting.
}

\section{Introduction}
\label{Sec:Intro}
In the recent advance of human assisting robot technologies, the ability to explain robot behaviors is becoming an important task.
Robot actions are often generated from uninterpretable systems, such as deep neural networks \cite{hatori2018interactively}.
However, robots are expected to be explainable for their behavior because they will cooperate with humans in daily life areas.
Most of the existing works for bridging robot actions and natural language captions are oriented to generated robot actions given natural language commands \cite{sugiura2011learning,kollar2010toward,tellex2011understanding,fasola2013using}.
They tackled the problem in a pipeline process by using manually designed base action units, i.e., small basic robot actions, which are easy to recognize and generate.
Both robot actions and captions were converted to sequences of base action units and then bridged each other.

In contrast, end-to-end approaches have been widely used to date by recent advances in the neural network field \cite{cho2014learning,sutskever2014sequence,chiu2018state}.
These approaches directly train mappings between two sequences.
Such end-to-end mapping is applied to robot behavior captioning, which generates a natural language caption given a raw robot action sequence \cite{takano2015statistical,plappert2018learning,yamada2018paired,thomason2019improving}.

It is not easy for human assisting robots to generate natural captions from their observations, sequences of actions, and observation from robot sensors, because robots encounter diverse environments and tasks.
Collecting large-scale dataset for any different robots or environment is not realistic.
Furthermore, one robot observation sequence has more samples than words in a caption due to sampling rates of observation devices.
However, this gap makes it challenging for neural networks to learn relationships between robot observations and natural language captions.
This problem is especially true if we apply neural network-based approaches, in particular recurrent neural networks, which are generally used for sequence-to-sequence learning.


Segmenting robot actions to base action units will contribute to making the training process more straightforward.
In the work we performed, we focused on using unsupervised segmentation methods \cite{nakamura2009grounding,nakamura2016continuous,nakamura2017segmenting} for speeding up and stabilizing sequence-to-sequence training, rather than using manually designed base robot action units \cite{fasola2013using}.
We used K-means clustering and chunking based on byte-pair encoding (BPE) to learn the unsupervised base action segments of the robot.
K-means clustering can unify similar robot observation into one class.
Chunking based on BPE can wrap a typical sequence of robot actions into one class. 

We also used an attention mechanism in sequence-to-sequence learning \cite{luong2015effective,vaswani2017attention}, which explicitly learns the correlation between input and output sequences.
Attention mechanism calculates correlation weights between each input sample and each output sample.
It is known that the attention mechanism can solve a problem of sample number gaps in the area of speech recognition and synthesis \cite{chorowski2015attention,wang2017tacotron}.
We expect that the training results obtained for the attention can be incorporated into a robot action unit vocabulary that corresponds to actual words in sentences.
We combined both methods; unsupervised action segment learning and attention mechanism, to generate more natural captions since we expected that each method would provide different contributions.


In experiments, we recorded videos of robots with their observation: actuator sequences and videos recorded by a first-person viewpoint camera.
We used these videos to collect robot behavior captions through crowdsourcing, which corresponds to robot action sequences.
However, we limited the number of collected samples toward a realistic situation to apply the method.
We conducted two experiments: an automatic evaluation based on references in the test-set and human subjective evaluation. 
These results showed that the proposed three types of base action units contribute to improving better robot behavior captions in natural language. 
In particular, robot action units acquired by the unsupervised learning based on clustering and chunking got the best performance in human subjective evaluation.


\begin{figure*}[t]
\centering
\includegraphics[width=\textwidth, clip]{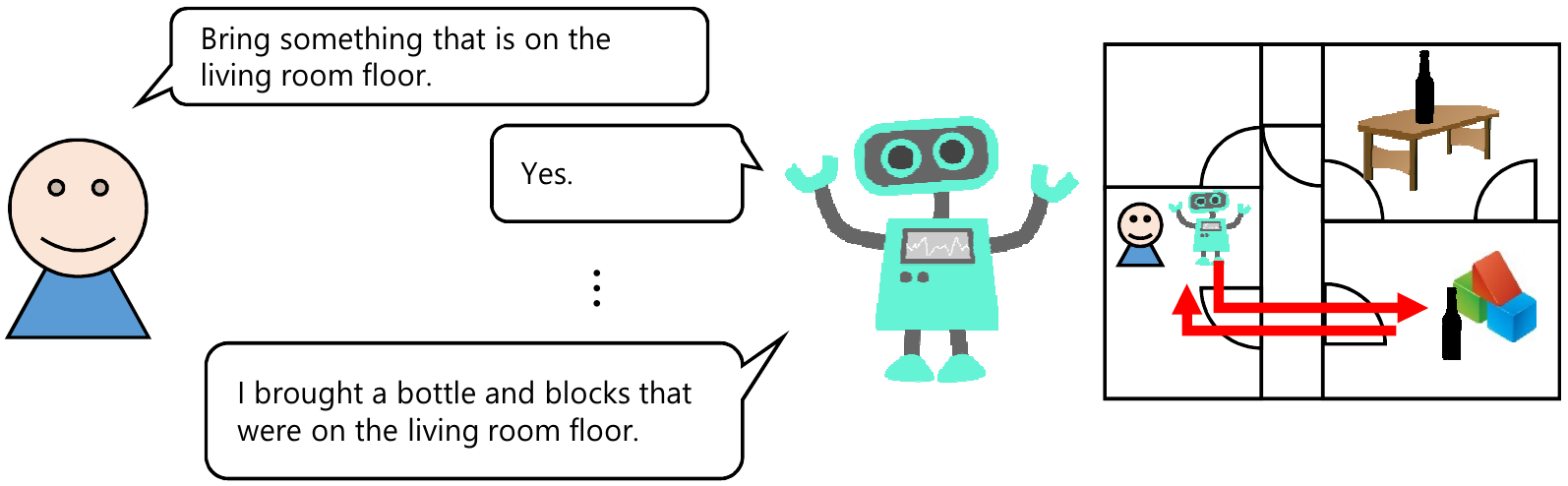}
\caption{Example human robot communication obtained through natural language commands/descriptions.}
\label{fig:example-dialog}
\end{figure*}

\section{Natural Language Captions for Robot Action Sequences}

\subsection{Problem Setting and Related Work}
For daily life scenarios, various robot tasks have been designed as human assisting robots have evolved.
In particular, these tasks include human assisting tasks, such as moving objects in rooms and recording videos of actual places.
In these situations, it is required for robots to communicate with users about their behavior in natural language.
There are two important directions. 
The first direction is deciding robot actions according to the natural language commands by the user. 
Another direction is providing robot behavior captions in natural language. 
Both directions are important to realize cooperative human assisting robots.
In these tasks, there are sequences of robot actions and natural language captions/descriptions, and the robot needs to learn the mapping between them.
Figure~\ref{fig:example-dialog} shows an example of the task in a particular situation.
In this example, the user requests the robot to ``Bring something that is on the living room floor,'' and the robot does so.
After this, the robot uses natural language to tell the user what was done, e.g., ``I brought a bottle and blocks that were on the living room floor''.
In this paper, we focus on the latter problem, generating a natural language caption given a robot action sequence.
We define the problem as generating a word sequence $Y$ given a sampled robot observation sequence $S$. 

As the robot action sequence $S$, most reported existing works have been done using the pipeline process, such as handcrafting base action units, recognizing all robot observation sequence portions with these base action units and constructing complicated actions \cite{kollar2010toward}.
However, these approaches for handcrafting base action units are costly.

Based on the advance of technologies based neural networks, some works tried to build sequence-to-sequence learning system between robot observations and natural language captions.
Takano et al. \cite{takano2015statistical} used bi-grams for end-to-end learning between natural language captions and robot action sequences, to separate the semantics and the language expressions in captions.
Yamada et al. \cite{yamada2018paired} utilized recursive autoencoders for the learning so as to minimize the distance between the latent variables of autoencoders for robot actions and natural language captions, toward the end-to-end bridging.
In another study, a simple encoder-decoder was applied to learn the relation \cite{plappert2018learning}.
However, these methods require large-scale training data.
It is not realistic to prepare such large-scale training data whenever the robot is applied to a new environment, or when the robot functions are updated.
The physical environments of robots themselves will be different if a new robot is developed.

To prevent problems caused by a lack of training data, in this paper, we applied two robot action segmentation types in an unsupervised manner to make it possible for the system to learn the relationship from a small data amount.
Some related studies have tried to define base action units with unsupervised learning \cite{nakamura2009grounding,nakamura2016continuous,nakamura2017segmenting} or adaptively \cite{iqbal2018toward}.
These studies focused on structuring robot actions from data.
Such clustering-based approaches have the potential to be used for reducing training data with simplifying symbols contained in training data.

\begin{figure}[t]
\centering
\includegraphics[width=70mm, clip]{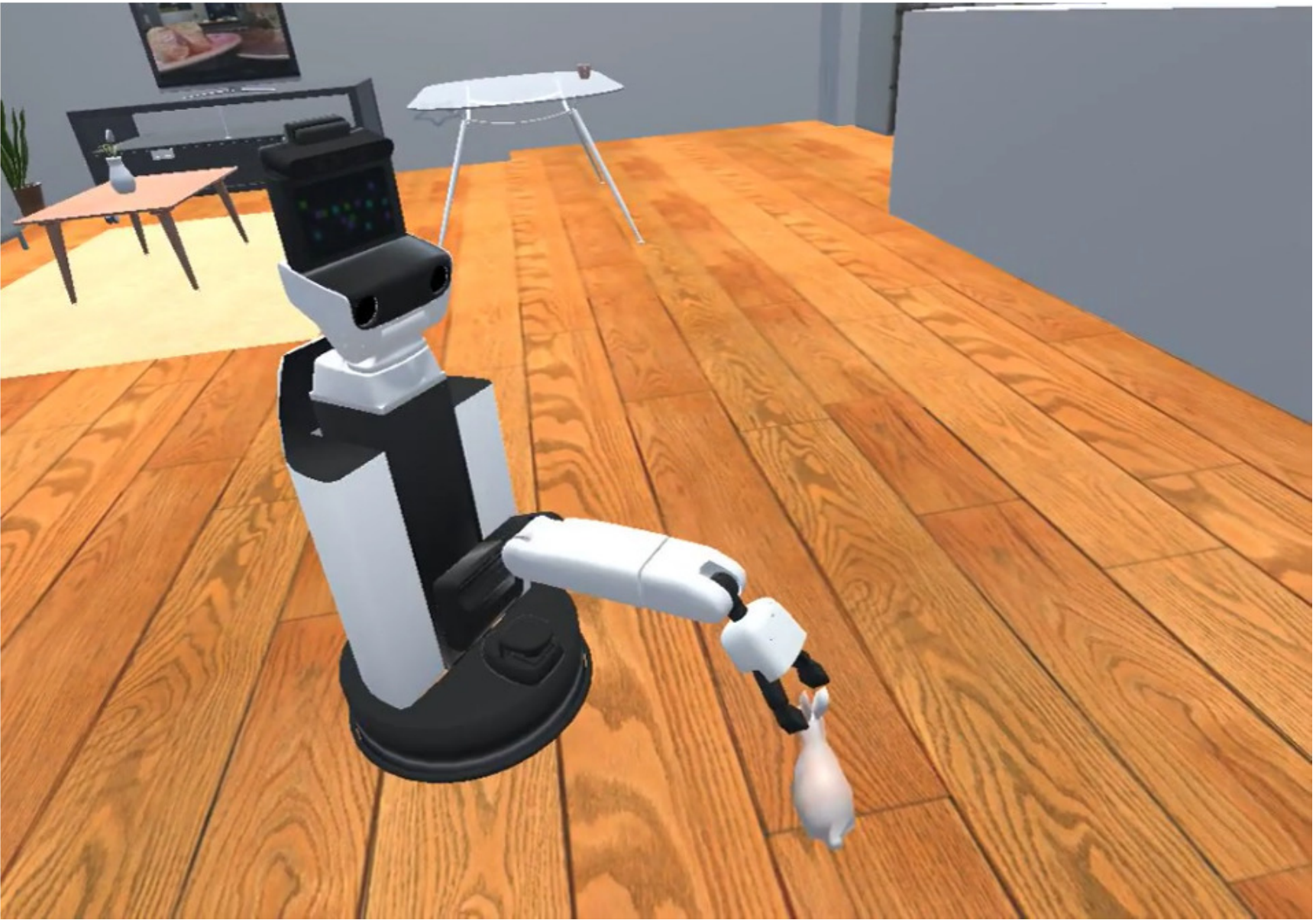}
\caption{HSR robot on SIGVerse environment, conducting a task in the setting of WRS.}
\label{fig:hsr}
\end{figure}

\subsection{Robot Task, Environment and Data in Simulator}
\label{sec:data}
In this section, we describe collected dataset for the training of captioning given robot observations.
We used the service category environment defined in the World Robot Summit (WRS) \cite{kimura2017competition} as the robot task for our action description generation.
We used Human Support Robot (HSR) \cite{yamaguchi2015hsr}, a human assisting robot in a domestic environment, and also used a robot simulator SIGVerse \cite{inamura2010simulator}.
We followed WRS Partner Robot Challenge (Virtual Space) instruction to build our environment.\footnote{https://worldrobotsummit.org/download/rulebook-en/rulebook-simulation$\_$league$\_$partner$\_$robot$\_$challnege.pdf}
The HSR robot in SIGVerse environment is shown in Figure~\ref{fig:hsr}.
The robot conducted tasks in this environment and described its actions with natural language behavior captions.
We generated robot actions in the simulator and added their captions\ky{, by using crowdsourcing,} as our training data. 

The input $S$ contains nine-dimensional motor actuation (rotations) and a three-dimensional robot moving (two directions and one rotation) that is sampled every 0.3 seconds.
The robot also uses 160$\times$120-pixel images from a first person viewpoint camera mounted on robot arm's tip.
The image feature is embedded into a ten-dimensional vector by using convolutional autoencoder (CAE) \cite{masci2011stacked}.
The robot action actuation and the image features are concatenated into one vector, which will be used as an input of encoder at each time-step.

We used crowdsourcing to add a caption $Y$ for a robot action sequence $S$.
We made videos of robot action sequences from third-person viewpoint camera as shown in Figure~\ref{fig:hsr}.
We showed the video to crowd-workers and then requested them to give their captions that describe behaviors of the robot in natural language.\footnote{Dataset including captions will be distributed when we publish the paper.}
We made 50 videos and added 20 captions for each.
Robot actions in the dataset contained ``bring'', ``put'', ``pick up'', ``drop'' and ``go to see'' actions and 10 different videos for each action.
Note that, we did not give crowd-workers any instructions on how they should include these verbs in descriptions. This means that, for example, they might use ``get'' as an alternative to ``bring'' in accordance with their language sense.
In total, we collected 1000 sentences in Japanese, which were segmented by KyTea \cite{neubig2011pointwise}.
Annotated captions are simple sentences, which consist of single predicate and several arguments, i.e., ``Drop the toy dog from the table'' or ``Pick up the teapot on the table''.

\subsection{Bridging based on Encoder-Decoder}

\begin{figure}[t]
\centering
\includegraphics[width=100mm, clip]{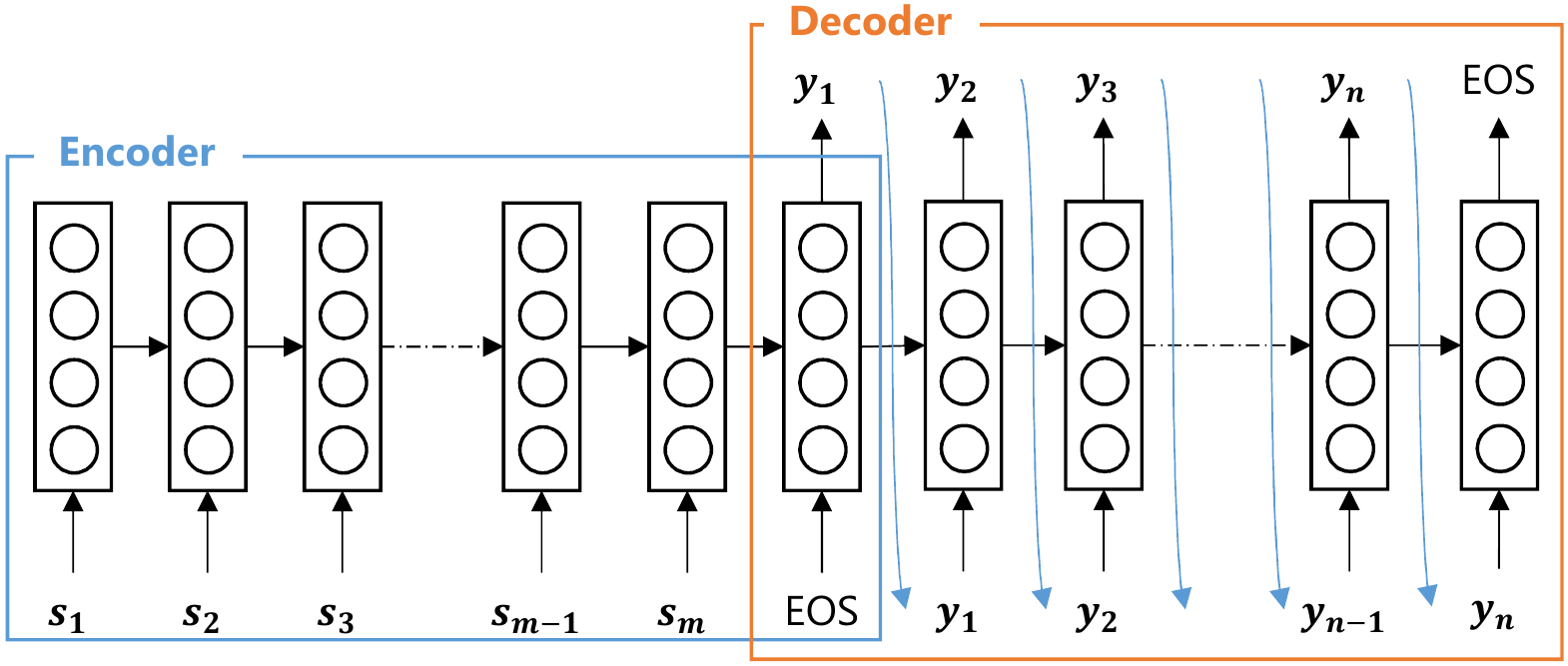}
\caption{Basic encoder-decoder architecture. In our problem, the encoder receives a sample of robot observation $s_i$ in time $i$ according to its sampling rate. Once the network receives the end-of-sequence symbol ({\sf EOS}), the decoder start to generate a description word sequence.}
\label{fig:encdec}
\end{figure}

We show the architecture of end-to-end relationship learning between the robot action sequence and the word sequence based on an encoder-decoder, which had been proposed previously \cite{plappert2018learning}.
The overall architecture in a time sequence is shown in Figure~\ref{fig:encdec}.
As mentioned in the problem setting, the system gets an input robot observation sequence $S$ to a generate natural language caption $Y$.
Here, $s_i$ indicates robot observations, the raw trajectory of robot action and a frame recorded by cameras on the robot, at time $i$ and $y_j$ indicates the $j$th word in the natural language description.
The encoder embeds the robot action $s_k$ in time $k$ to the hidden layer $h_k$ as
\begin{align}
h_i &= \sigma (W_{sh}s_i + W_{hh}h_{i-1} + b_h),
\end{align}
where, $W_{sh}$ and $W_{hh}$ are weight matrices for conversion and $b_h$ is a bias.
$\sigma$ is the activation function.
The decoder starts generating word $w_j$ as soon as the encoder receives the end-of-sequence ({\sf EOS}) symbol and continues to generate as it generates $w_{n+1}$.
The network is updated by
\begin{align}
h_i &= \sigma (W_{yh}w_{i-1} + W_{hh}h_{i-1} + b_h),\\
y_i &= softmax (W_{hy}h_i + b_y),
\end{align}
where $W_{yh}$, $W_{hh}$ and $W_{hy}$ are decoder weight matrices, and $b_h$ and $b_y$ are bias.
$softmax$ is the activation function to predict words.
This architecture was proposed in the task of machine translation and is widely used on a variety of mapping tasks between sequences, such as conversational response generation.
The difference from machine translation or conversational response generation tasks is that the robot action sequence $S$ will be very long in accordance with the robot trajectory sampling rates.
This property increases the risk of that vanishing gradients \cite{bengio1994learning}.
In the area of speech recognition or synthesis, attention mechanism is used to solve the problem \cite{chorowski2015attention,wang2017tacotron}.
The problem can be mitigated if there is sufficient training data; however, this means that large-scale new data will be required for each new domain or each robot task.

\section{Robot Action Segmentation}
\label{sec:methods}

The dataset we used had only 1000 samples, corresponding to 50 actions, thus making it difficult to learn a good mapping with a vanilla encoder-decoder.
However, as we mentioned before, collecting a large-scale dataset for all domains and robots is not a realistic option.
Thus, we trained robot action segments as base action units by using unsupervised learning: clustering and chunking.
We also incorporated the attention mechanism into the encoder-decoder, since we expect the attention weights could learn the action classes.

\subsection{Robot Action Segmentation Based on Clustering and Chunking}

\begin{figure}[h]
\centering
\includegraphics[width=60mm, clip]{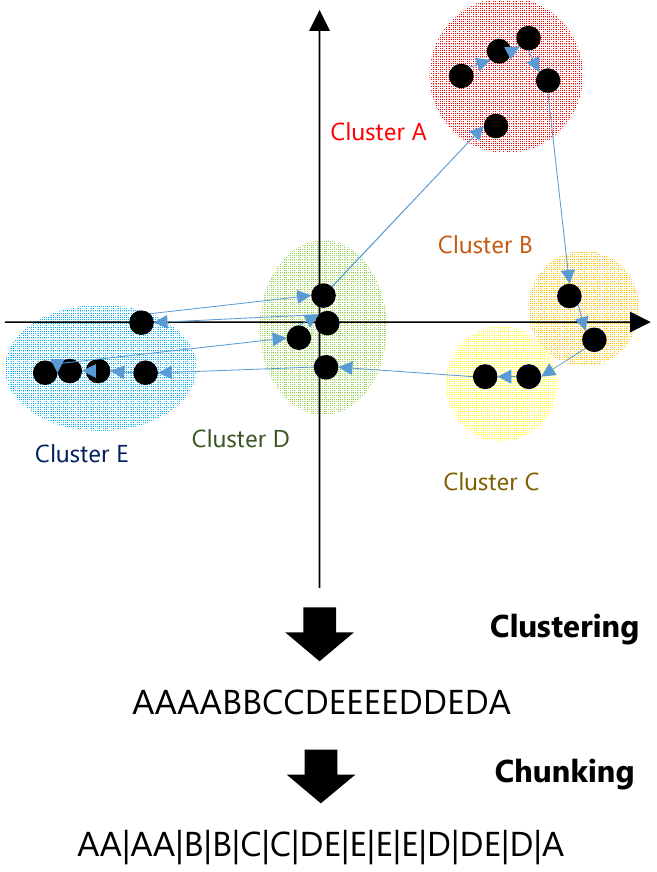}
\caption{Robot action segmentation process: clustering and chunking.}
\label{fig:clustering-chunking}
\end{figure}

Our system samples raw robot actions every 0.3 seconds. 
Figure~\ref{fig:clustering-chunking} showing a clustering and chunking example.
This graph shows robot actions in only two dimensions for an easy description.
Robot action vectors are quantized by k-means clustering \cite{hartigan1979algorithm}, which decides what the centroids should be for the defined number of classes.
In this example, the number of samples is 18 and the number of class is set as 5.
We used the Elbow method \cite{bholowalia2014ebk} to determine the class number should be 150.

After the clustering, we extracted some chunks by using byte-pair encoding (BPE).
BPE extracts frequent symbol sequences and replace them to another symbol as maximizing the compression of a given raw sequence.
The robot action sequence quantized by clustering is still longer than the word sequence; however, we can shorten the sequence by applying BPE.
The BPE units can be used as base action units. 
In the Figure~\ref{fig:clustering-chunking} example, {\sf AA} and {\sf DE} are extracted as vocabularies.
In our experiments, we set the BPE vocabulary size as 200 and trained the vocabulary from the training data.
We refer to this method as ``explicit segmentation'' in the following sections.

\begin{figure*}[t]
\centering
\includegraphics[width=100mm, clip]{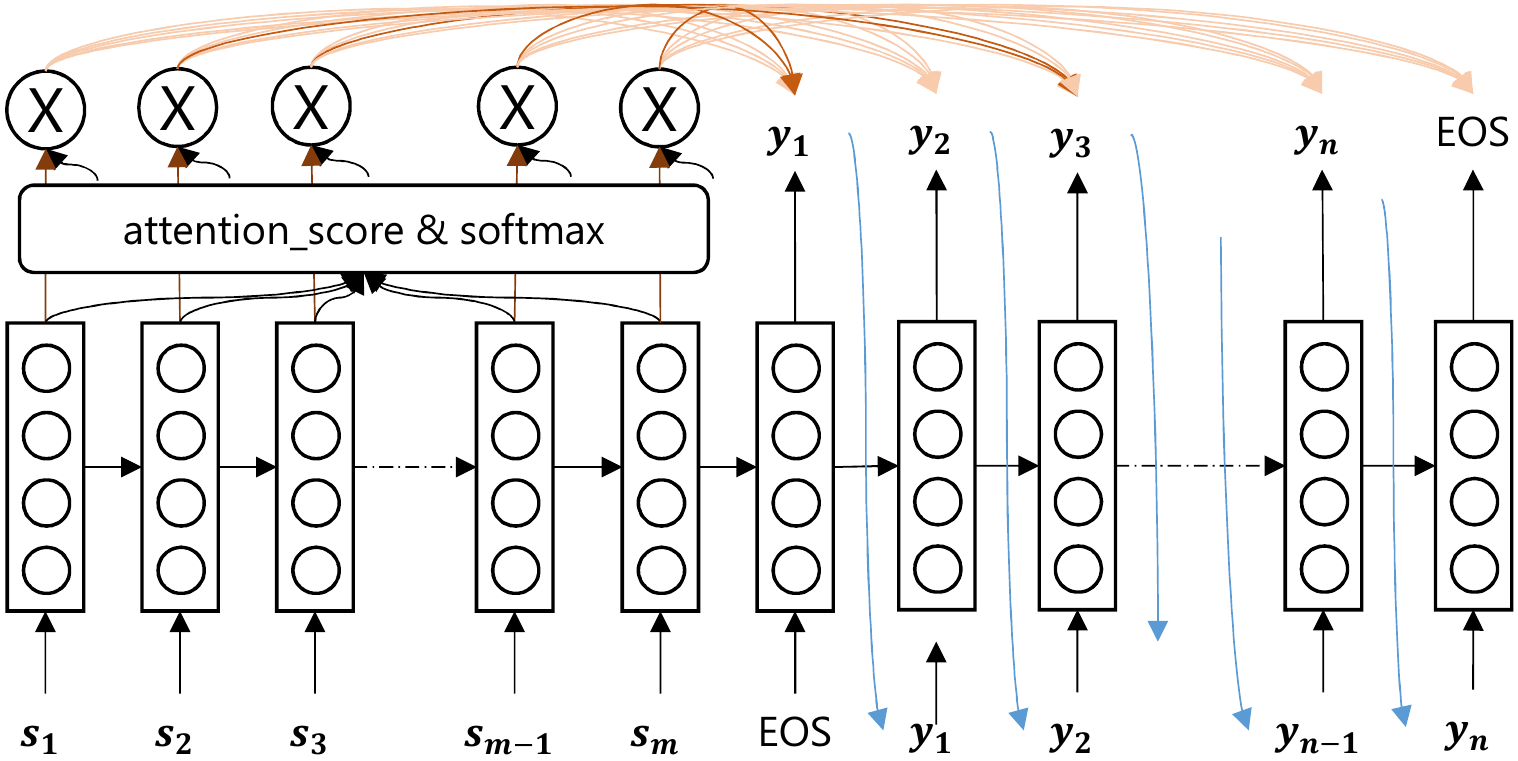}
\caption{Encoder-decoder with attention mechanism.}
\label{fig:attention-mechanism}
\end{figure*}

\subsection{Segmentation Based on Attention Mechanism}

We conducted explicit segmentation as a pre-processing of relation learning by using encoder-decoder neural networks.
In contrast, the attention mechanism can consider input chunks in attention weight learning implicitly.
The attention mechanism learns the mapping from an input sample to an output sample by using a gating mechanism.
In our situation, the number of input samples is larger than the number of output samples; thus input samples associated with the same output sample can be interpreted that they belong to the same class.

Attention mapping is calculated between the hidden layer of encoder $h_{e,i}$ and the hidden layer of decoder $h_{d,j}$ as,
\begin{align}
a_{i,j} &= h_{e,i}^{T}W_{a}h_{d,j}.
\end{align}
An attention weight is trained as a part attention vector $a_j$ in each decoding step.
Input samples that have similar attention weights for an actual output sample are strongly used in the decoding of the associated output.
We expect that this property of attention mechanism works as segmentation.
We show an encoder-decoder that has an attention mechanism in Figure~\ref{fig:attention-mechanism}.
In this example, strong attention weights are indicated in a deep color. 
States $s_2$ and $s_3$ have higher attention weights to $y_3$ in the output, and states $s_{m-1}$ and $s_m$ have higher attention weights to $y_1$. 
We expect that such weights in the attention mechanism work as base action units.
We hereafter call this method ``implicit segmentation''.

Note that, it is known that the attention mechanism contributes to improve the entropy of the generated sentence; however, it also often causes an overfitting and dull generations. In other words, sentence naturalness will be improved by the attention mechanism; however, the system often generate similar sentences to different input sequence.

\subsection{Hybrid Method of Implicit and Explicit}

Explicit segmentation segments sequences by focusing on input sequence information, in contrast to that of implicit segmentation, which focuses on the relationship between the input sequence and the output sequence for segmenting the input sequence.
In other words, these segmentations use different types of information.
Thus, we integrated these segmentations into a single method as ``hybrid segmentation'' to make the best of them both.
In the hybrid segmentation, we used explicit segmentation to segment the data as pre-processing, and then fed the data into the encoder-decoder with the attention mechanism to combine with the implicit segmentation method.

\section{Experiments}
In our experiments, we used the data collected in Section~\ref{sec:data} to train an encoder-decoder that generates a word sequence of natural language caption $T$ given a raw robot action sequence $S$.
As the baseline, we prepared a vanilla encoder-decoder, because the focus of this paper is investigating a good robot action segmentation, which makes it possible to train the encoder-decoder from a small amount of data.
We compared three methods described in Section~\ref{sec:methods}; explicit segmentation, implicit segmentation, and hybrid segmentation, with the vanilla encoder-decoder.
Experimental details follow.

\subsection{Experimental Settings}
We separated our 50 robot actions into 40/5/5 as training/validation/testing datasets and conducted ten cross-fold validation to evaluate all the data.
The number of decriptions associated with one robot action was 20; thus training/validation/testing datasets cotains 800/100/100 pairs of sequences.
We used a one-layered LSTM that has 160 units as our encoder-decoder model.
The batch size was 64, the dropout rate was 0.5, the learning rate was 0.001, and the weight decay was 1e-0.6.
We decided the number of training epochs by using the validation dataset.

\begin{table}[t]
\centering{
\caption{Evaluation with BLEU scores. \ky{Note that the range is from 0 to 100.}\label{tab:bleu}}
  \begin{tabular}{l||r|r|r}
    Model & BLEU-2 & BLEU-3 & BLEU-4\\
    \hline \hline
    Vanilla & 0.1 & 0.0 & 0.0\\ \hline
    Explicit & 18.0 & 11.7 & 8.4\\ \hline
    Implicit & 21.4 & 14.7 & 10.4\\ \hline
    Hybrid & 23.4 & 16.2 & 11.7\\ 
  \end{tabular}
  }
\end{table}

\subsection{Automatic Evaluation with BLEU}
We used automatic evaluation metrics BLEU-2, 3 and 4 \cite{papineni2002bleu} to evaluate the generated captions, by using references annotated to videos in the testing set.
BLEU-$n$ calculates the ratio of matched $n$-grams with smaller numbers of $n$, between the genereation and the reference.
Each video has 20 reference sentences; thus we used multi-reference setting in the automatic evaluation. 
We calculated BLEU-$n$ scores for each reference and selected a reference that has the best BLEU-$n$ score for the evaluation.

Evaluation scores, BLEU-2, 3 and 4, for each model are shown in Table~\ref{tab:bleu}.
Vanilla encoder-decoder scores were deficient and scores are improved by introducing segmentation methods.
Both the explicit segmentation and the implicit segmentation were effective in improving scores.
The hybrid segmentation achieved the best score in each metric; it indicates the explicit segmentation and the implicit segmentation made different contributions to each other.
However, scores are slightly low even if the task is simple.
We will investigate the reason on the next human subjective evaluation and generation analysis.



\begin{table*}[t]
\centering{
\caption{Subjective evaluation result distribution for each method.\label{tab:subjective}}
  \begin{tabular}{l||r|r|r|r|r||r}
    Model & a & b & c & d & e & a--c\\
    \hline \hline
    Vanilla & 0.0\% & 0.0\% & 0.0\% & 0.0\% & {\bf 100.0}\% & 0.0\%\\ \hline
    Explicit & {\bf 10.6}\% & {\bf 17.2}\% & {\bf 37.2}\% & 28.3\% & 6.7\% & {\bf 65.0}\%\\ \hline
    Implicit & 3.9\% & 5.0\% & 35.0\% & {\bf 56.1}\% & 0.0\% & 43.9\%\\ \hline
    Hybrid & 7.2\% & 8.9\% & 33.3\% & 43.9\% & 6.7\% & 49.4\%\\ 
  \end{tabular}
  }
\end{table*}

\begin{table*}[t]
\centering{
\scriptsize{
\caption{Generation examples. The ``score distribution'' indicates numbers of evaluators who gave a, b, c, d, and e. \label{tab:example}}
  \begin{tabular}{l||l|c}
  Method & Generation results (Translation) & Score distribution\\ \hline \hline
  Reference & inu no omocha wo tehburu kara otoshite & - \\
            & (Drop the toy dog from the table.)   &  \\
  Vanilla   & te te te te te te te te te te te te te (Repeating function words.)   &  e:9 \\
  Explicit  & ringo wo otoshite (Drop the apple.)   & b:2, c:7  \\
  Implicit  & tehburu no ue no sohsu wo totte kite (Bring the sauce on the table.) & d:9  \\
  Hybrid    & tehburu no ue no koppu wo totte kite (Bring the cup on the table.)   & d:9  \\ \hline
  Reference & tehburu no ue wo nozoki konde (Look at the top of the table.)   & - \\
  Vanilla   & te te te te te te te no no no no no no (Repeating function words.)   & e:9  \\
  Explicit  & shinshitsu no yousu wo mite (Look at the bed room.)   & b:3, c:5, d:1  \\
  Implicit  & tehburu no ue no sohsu wo totte kite (Bring the sause on the table.)   & a:1, d:7, e:1  \\
  Hybrid    & heya no ue no yousu wo totte (Bring the state of the upper of the room.)   & d:2, e:7  \\ \hline
  Reference & yuka no thipotto wo hirotte (Pick up the teapot on the floor.)   & - \\
  Vanilla   & te te te te te te te te te te te te te (Repeating function words.)   & e:9 \\
  Explicit  & yuka no sohsu wo totte (Pick up the sauce on the floor.)   & a:7, b:1, c:1  \\
  Implicit  & tehburu no ue no sohsu wo totte (Pick up the sause on the table.)  & b:1, c:7, d:1 \\
  Hybrid    & yuka ni aru sohsu wo totte (Pick up the sauce placed on the floor.)   & a:7, c:2 \\ \hline
  Reference & tehburu no koppu wo katadukete (Clear away the cup on the table.)   & - \\
  Vanilla   & te te te te te te te te te te te te te (Repeating function words.)   & e:9  \\
  Explicit  & tehburu no ue no okimono wo totte kite (Bring the ornament on the table.)   & a:1, b:5, c:3 \\
  Implicit  & tehburu no ue no sohsu wo totte kite (Bring the sause on the table.) & b:1, c:8 \\
  Hybrid    & tehburu no ue no okimono wo totte kite (Bring the ornament on the table.)   & a:1, b:4, c:3, d:1  \\ \hline
  Reference & tehburu no thipotto wo kauntah ni oite  & - \\
            & (Bring the teapot from the counter to the table) & \\
  Vanilla   & te te te te te te te no no no no no no (Repeating function words.)   & e:9  \\
  Explicit  & kicchin no ue ni aru kan wo totte kite (Bring the can on the kitchen.)& b:1, c:3, d:5\\
  Implicit  & tehburu no ue no sohsu wo totte kite (Bring the sause on the table.)   & a:6, b:3  \\
  Hybrid    & kicchin no ue ni aru ka mite kite (Look at the upper of kitchen to check.)   & d:9  \\ \hline
  \end{tabular}
}  }
\end{table*}
\subsection{Human Subjective Evaluation}
The automatic evaluation with BLEU calculates distance from references to generated sentences.
The results have moderate correlations to the quality of generation; however, there is still a gap between the result of evaluation with BLEU and human impression to the quality.
For example, it is challenging to evaluate variations of word selection by only using BLEU.
Thus, we conducted a human subjective evaluation to investigate the qualities of generated sentences from each method.

We showed both robot action videos and their associated captions generated by models to human subjects for scores these generated captions.
In the experiment, each subject evaluated 20 sentences for each method, 80 sentences in total (for 4 methods).
Following evaluation criteria are shown to human subjects.
\begin{enumerate}
\renewcommand{\labelenumi}{\alph{enumi}).}
  \setlength{\parskip}{0cm} 
  \setlength{\itemsep}{0cm} 
    \item The sentence is exactly describing the robot action in the video. 
    \item The sentence is mostly describing the robot action in the video; even it contains some minor errors. 
    \item The sentence is moderately describing the robot action, but the description still has some major errors on target objects or environments.
    \item The sentence is grammatically correct; however, the action name and the object name are different.
    \item The sentence has wrong grammar and is meaningless.
\end{enumerate}
a)--c) are prepared to check semantics of generated captions in different levels, compared with d) and e) focus more on sentence syntax.
Human subjects selected one of them for the presented pair of a robot action video and its generated captions. 
The number of subjects was 9; the total number of evaluated example for each method was 180.
Table~\ref{tab:subjective} shows results.

From the results, we can know that the vanilla encoder-decoder did not generate any meaningful sentences; this result indicates that it is challenging to learn the mapping between robot action sequences and caption word sequences from a small portion of data.
In contrast, tried segmentation methods make it possible the model to generate many meaningful sentences; in particular, the explicit segmentation method generated meaningful sentence a lot.
Human subjects judged that 65\% of generated sentences from the explicit segmentation method is meaningful (a, b and c); this result indicates the usefulness of our proposed segmentation method.
Overall score is still slightly poor because of the number of training samples, as a result, the majority in the evaluation result was c.
This result indicates that the explicit segmentation method successfully learned predicates from robot action trajectory; however, failed to learn about objects names from camera observations in most cases.
The implicit segmentation method also improved the score from the vanilla system score; however, the ratio of meaningful generated sentences was 43.9\%; it was less than the result of the explicit segmentation method.
This result indicates that it is challenging to learn the relation by using only the attention mechanism, even if the effect is high.
The implicit segmentation method generated no ungrammatical sentence; this result indicates the potential of using attention mechanism, even if its score is less than the explicit segmentation method.
This is probably because the attention mechanism is more focusing on fluency than content words.
The score of the hybrid segmentation method was slightly less than the explicit segmentation method, even if the method achieved the best BLEU score in the automatic evaluation.
This problem is probably caused by overfitting of word prediction in the softmax cross entropy loss calculation of encoder-decoder.
However, the explicit segmentation method lightened the problem of overfitting than other methods.
In summary, the explicit segmentation method, which uses clustering and chunking, achieved the best score for captioning robot actions in an end-to-end manner, in particular, generating a caption of actions (predicates) from the robot action trajectory.
However, we still have a problem with generation details, i.e., object names in descriptions.


\subsection{Discussing Generated Sentences}
Table~\ref{tab:example} shows examples generated from each method with their scores on human subjective evaluation.
First, it can be seen that the vanilla method generated ungrammatical sentences in most cases.
This supports our assumption that it is tough for a vanilla encoder-decoder to train the mapping from a small data-set.
In contrast, the explicit segmentation method generated grammatical sentences.
In particular, verbs in sentences were correct in most cases. 
This indicates that our method successfully learned the mapping between robot trajectories and verbs.

Generations from the implicit segmentation method was fluent; however, it sometimes contain similar sentences even if robot actions are different.
For example, if we look at the first and second examples, only the explicit segmentation method can generate verbs ``drop'' and ``look'', compared with implicit and hybrid generate ``bring'', which is the most frequent verb in our training data.
As observed elements, the implicit segmentation method often generated the word ``sauce'', because the word was frequent in the training data.
This is probably caused by the overfitting of neural networks and generating dull generations in many cases.
The explicit segmentation method reduced the problem in some cases.

The other problem that indicates the overfitting of the system is generating ``sauce'' and ``table'' as objects.
The implicit segmentation method generated these words frequently than other methods.
These words are more frequent than other objects; however, the implicit method generated these words in most cases, even if the other two methods generated different object and place names.
This indicates the difficulty of our task, learning a mapping between robot observations and natural language captions from a small amount of data, and the contribution of clustering and chunking with explicit methods.

In many cases, robots do not successfully identify objects.
This is probably since the system misunderstood visual information because our method did not have an object recognition module.
Pre-training based on object labels is a possible solution solve this problem.

\section{Summary}
\label{Sec:Conc}

This paper presented a method that generates robot behavior captions in natural language setting from robot action sequences, trajectories, and images, toward a cooperative human assisting robot that has explainability.
The method uses two different segmentation types for robot action actuation for training from small-size data: explicit segmentation based on clustering and chunking and implicit segmentation based on the attention mechanism of neural networks.
We also described the hybrid segmentation that integrates both types.
Experimental results indicated that the proposed methods improved generated caption quality, especially for generating verbs related to the robot trajectories.

We are still trying to improve the method's ability to describe objects, and so we recognize that a subject for future works will be to use an object recognition system as a module.
We applied simple clustering and chunking methods; thus, other segmentation methods or different types of robot action input to the network (e.g., VideoBERT \cite{sun2019videobert}) have the potential to improve scores.
Unsupervised learning on generated data on the simulator is also a prospective approach to improve scores.
Another subject for future work will be to apply our method to generate robot action sequences when it receives natural language commands from users.

\section*{Acknowledgement}
Part of this work was supported by JSPS KAKENHI Grant Number JP17H06101.


\end{document}